\documentclass[10pt, a4paper]{article}

\pdfoutput=1
\usepackage{amsmath}
\usepackage{authblk} % for affiliations
	\renewcommand\Affilfont{\itshape\footnotesize}
\usepackage{natbib}
\bibpunct{(}{)}{,}{a}{}{,}  % to adjust punctuation in references
\usepackage{graphicx} 
\usepackage{hyperref}
	\hypersetup{colorlinks=FALSE, breaklinks=true, linkcolor=darkblue, menucolor=darkblue, urlcolor=darkblue, citecolor=darkblue, hidelinks=TRUE}

\usepackage{multicol}              
\usepackage{multirow}
\usepackage{booktabs} 
\usepackage{geometry}
\usepackage[UKenglish]{isodate}
	\cleanlookdateon

\usepackage{color}
\usepackage{xcolor}
	\definecolor{darkblue}{rgb}{0,0,.5}
	\definecolor{darkgreen}{rgb}{0.0, 0.392, 0}
	\definecolor{darkgrey}{rgb}{0.4,0.4,0.4}
	\definecolor{grey}{rgb}{0.7,0.7,0.7}
	\definecolor{lightblue}{rgb}{0.678,0.847,0.902}
	\definecolor{blue}{rgb}{0,0,1}
	\definecolor{darkblue}{rgb}{0,0,0.545}
	\definecolor{green3}{rgb}{0,0.804,0.0}
\usepackage{mathptmx}
\usepackage{lineno}
\usepackage[margin=10pt, font=small, labelfont=bf, tableposition=top]{caption} 
\newcommand*{\xdash}[1][1em]{\rule[0.5ex]{#1}{1.3pt}}

\usepackage{tikz}
\newcommand{\tikzcircle}[2][red,fill=red]{\tikz[baseline=-0.5ex]\draw[#1,radius=#2] (0,0) circle ;}

\author[1]{Severin Hauenstein} %\thanks{severin.hauenstein@saturn.uni-freiburg.de}
\author[1]{Carsten F. Dormann\thanks{carsten.dormann@biom.uni-freiburg.de}}
\author[2]{Simon N. Wood}%\thanks{simon.wood@bristol.ac.uk}
\affil[1]{Department of Biometry and Environmental System Analysis, University of Freiburg, 79106 Freiburg, Germany}
\affil[2]{School of Mathematics, University of Bristol, Bristol BS8 1TW, U.K.}
\title{Computing AIC for black-box models using Generalised Degrees of Freedom: a comparison with cross-validation}
\begin{document}
\maketitle
\begin{abstract}
%to be submitted to (for example) Journal of Agricultural, Biological, and Environmental Statistics (wants LaTeX style from here: http://www.biometrics.tibs.org/latex.html); Environmental and Ecological Statistics; Statistics and Computing; Journal of Computational and Graphical Statistics; Computational Statistics and Data Analysis; Open Journal of Statistics
%
\noindent Generalised Degrees of Freedom (GDF), as defined by Ye (1998 JASA 93:120-131), represent the sensitivity of model fits to perturbations of the data. As such they can be computed for any statistical model, making it possible, in principle, to derive the number of parameters in machine-learning approaches. Defined originally for normally distributed data only, we here investigate the potential of this approach for Bernoulli-data. GDF-values for models of simulated and real data are compared to model complexity-estimates from cross-validation. Similarly, we computed GDF-based AICc for randomForest, neural networks and boosted regression trees and demonstrated its similarity to cross-validation. GDF-estimates for binary data were unstable and inconsistently sensitive to the number of data points perturbed simultaneously, while at the same time being extremely computer-intensive in their calculation. Repeated 10-fold cross-validation was more robust, based on fewer assumptions and faster to compute. Our findings suggest that the GDF-approach does not readily transfer to Bernoulli data and a wider range of regression approaches.
\end{abstract}
%\pagebreak

%\linenumbers

% ------------------------------------------------------------ %
\section{Introduction}
% Scene setting
%1st approach:
In many scientific fields, statistical models have become a frequently used tool to approach research questions. Choosing the most appropriate model(s), i.e. the model(s) best supported by the data, however, can be difficult. Especially in ecological, sociological and psychological research, where data are often sparse while systems are complex, the evidence one particular statistical model may not be exclusive. The existence of alternative models, which fit the data with comparable goodness, but yield considerably different predictions is ubiquitous \citep[e.g.][]{Draper1995,Madigan1995,Raftery1996}.

% This uncertainty can be decreased by averaging several models, weighted by their relative goodness \citep[e.g.][]{Hoeting1999,Vrugt2007}, e.g. such that
%\begin{equation}
%\label{ensemble}
%\overline{\widehat{y}}=\sum_{m=1}^{n}w_m\widehat{y}_m,
%\end{equation}
% how about a citation for this equation? 
%where $w_m$ represents a weighting factor of the predictions $\widehat{y}_m$ made by model $m$. 

%Both single model selection and model averaging require a valid method to assess the quality of the candidate models.
%In the late 1990s, information-theoretical approaches were brought to the applied statistical domain by the seminal publication of \citeauthor{Burnham1998} \citeyearpar{Burnham1998,Burnham2002}. 
In ecological and evolutionary research, statistical procedures are dominated by an ``information-theoretical approach'' \citep{Burnham1998,Burnham2002}, which essentially means that model fit is assessed by Akaike's Information Criterion \citep[defined as $\text{AIC}=-2\text{ log-likelihood}+2\text{ number of parameters}$:][]{Akaike1973}. In this field, the AIC has become the paradigmatic standard for model selection of likelihood-based models as well as for determination of model weights in model averaging \citep[e.g.][]{Diniz-Filho2008,Mundry2011,Hegyi2011}. %Since then, model averaging and selection based on AIC (defined as $\text{AIC}=-2\text{log-likelihood}+2p$) have been used in hundreds of studies.
%2nd approach:
%The use of statistical models has become a standard technique in many scientific disciplines. Occasionally, however, their application is compromised, essentially due to insufficient statistical expertise. In the late 1990s, information-theoretical approaches were brought to the ecological domain by the seminal publication by \citeauthor{Burnham1998} \citeyearpar{Burnham1998,Burnham2002}. Since then, model averaging and model selection based on AIC (defined as $\text{AIC}=-2\text{log-likelihood}+2p$) have been used in many studies.
%AIC is defined as $\text{AIC}=-2\ell_m+2p$, where $\ell_m$ is the maximised log-likelihood of model $m$, and $p$ the number of parameters \citep{Akaike1973}. %It follows the idea of parsimony and balances the goodness of fit of the model and its complexity for which we assume the number of fitted model parameters $p$ to be a valid measure.
% Problem
%With the wave of machine-learning methods in the early 2000s \citep[e.g.][]{Recknagel2001,Olden2008}, however, it has become even more difficult to select between, or average over, models. 
In recent years, and particularly in the field of species distribution analyses, non-parametric, likelihood-free approaches (`machine learning')  have become more prevalent and in comparisons typically show better predictive ability \citep[e.g.][]{Recknagel2001,Elith2006,Olden2008,Elith2009}. However, these approaches do not allow the computation of an AIC, because many of such `black-box' methods are neither likelihood-based, nor can one readily account for model complexity, as the number of parameters does not reflect the effective degrees of freedom (a phenomenon \citet{ElderIV2003} calls the ``paradox of ensembles"). Thus, at the moment ecologists are faced with the dilemma of either following the AIC-paradigm, which essentially limits their toolbox to GLMs and GAMs, or use machine-learning tools and closing the AIC-door. From a statistical point of view, dropping an AIC-based approach to model selection and averaging is no loss, as an alternative approximation of the Kullback-Leibler distance is possible through cross-validation. 

%Using some sort of AIC seems reasonable, on the basis that it is not obviously worse than the alternatives and AIC is often simple to compute, provided we have a well defined notion of effective degrees of freedom. However the latter is often problematic.

% Our approach
In this study, we explore a potential avenue to unite AIC and machine learning, based on the concept of Generalised Degrees of Freedom (GDF). We explore the computation of GDFs for Gaussian and Bernoulli-distributed data as plug-in estimates of the number of parameters in a model. We also compare such a GDF-AIC-based approach with cross-validation.
 
The paper is organised as follows. We first review the Generalised Degrees of Freedom concept and relate it to the degrees of freedom in linear models. Next, we briefly illustrate the relation between cross-validation and Kullback-Leibler divergence, as KL also underlies the derivation of the AIC. Through simulation and real data we then explore the computation of GDFs for a variety of modelling algorithms and its stability. The paper closes with a comparison of GDF-based AIC and cross-validation-derived deviance, and comments on computational efficiency of the different approaches and consequences for model averaging.

%Here we present two approaches to quantify the goodness-of-fit of a model. Our first approach is to compute AIC-values for black-box models using Generalised Degrees of Freedom (GDF) \citep{Ye1998}. This allows us to align machine-learning approaches with GLM-like parametric models, making them accessible to the same information-theoretical approach for model selection and averaging. The key problem we address here is how to compute the effective degrees of freedom in a black-box algorithm, i.e. in a model that internally resorts to neural networks, recursive partitioning, boosting, bagging, randomisations, regularisation, cross-validation and/or various kinds of optimisation routines. Even if it was statistically possible to analytically keep track of the effect of \emph{each} such refinement on the number of parameters of the model, in combination this becomes infeasible. 

%In a second approach we use repeated 10-fold cross-validation to assess the strength of a model.

%In the following, we briefly outline the theory behind the two approaches and how we adapt GDF to binary data. In the analysis we first test how GDF performs for several configurations. We then provide a simple model selection and averaging scenario on two simulated and two real-life datasets and compare the results of both approaches.   

\subsection{Generalised Degrees of Freedom}
% Introducing GDF
Generalised Degrees of Freedom (GDF), originally proposed by \citet{Ye1998} and illustrated for machine learning by \citet*{ElderIV2003}, can be used as a measure of model complexity through which we can make information theoretical approaches applicable to blackbox algorithms. 

In order to understand the conceptual properties of this method, we can make use of a somewhat simpler version of degrees of freedom (df).
For a linear model  \emph{m}, $ \mathbf{y=X\boldsymbol\beta+\boldsymbol\epsilon}$, df are computed as the trace of the hat (or projection) matrix \textbf{H}, with elements $h_{ij}$ \citep[e.g.][p.~153]{Hastie2009}: 
\begin{equation}
\text{df}_m=\mathrm{trace}\left(\mathbf{H}\right)=\mathrm{trace}\left(\mathbf{X}\left(\mathbf{X}^T \mathbf{X}\right)^{-1} \mathbf{X}^T\right).
\end{equation}
For a linear model this is the number of its independent parameters, i.e. the rank of the model. To expand this concept to other, non-parametric methods, making it in fact independent of the actual fitted model, the definition above provides the basis for a \emph{generalised} version of degrees of freedom. Thus, and according to \citet{Ye1998}, 
\begin{equation}
\mathrm{GDF}_m = \mathrm{trace}\left(\mathbf{H}\right) = \sum_{i}^{}h_{ii} = \sum_{i}^{}\frac{\partial\hat{y}_i}{\partial y_i}, \label{eqn:GDF1}
\end{equation}
where $\hat{y}_i$ is the fitted value. That is to say, a model is considered the more complex the more adaptively it fits the data, which is reflected in higher GDF. Or in the words of \citet[p.~120]{Ye1998}: GDF quantifies the ``sensitivity of each fitted value to perturbations in the corresponding observed value''. For additive error models (i.e. $Y = f(X) + \varepsilon$, with $var(\varepsilon) = \sigma_\varepsilon^2$), we can use the specific definition that 
\begin{equation}
\mathrm{GDF}_m = \frac{\sum^N_{i=1}cov(\hat{y}_i), y_i}{\sigma_\varepsilon^2}
\end{equation}
\citep[][p.~233]{Hastie2009}. Beyond the additive-error model more generally, however, we have to approximate eqn~\ref{eqn:GDF1} in other ways. 
By fitting the model with perturbed $y_i$ and assessing the response in $\hat{y}_i$ we can evaluate $\frac{\partial\hat{y}_i}{\partial y_i}$ so that,
\begin{equation}
\label{gdf_vert}
\mathrm{GDF}\approx \sum_{i}^{}\frac{\widehat{y'}_i-\hat{y}_i}{y'_i-y_i},
\end{equation}
where $y'_i$ is perturbed in such a way that, for normally distributed data, $y'_i=y_i+\operatorname{N}\left(\mu=0,\sigma\right)$ with $\sigma$ being  small relatively to the variance of $y$. 
%For the computation of GDF \citet{ElderIV2003} proposed to perturb all $y_i$ simultaneously.

In order to adapt GDF to binary data we need to reconsider this procedure. Since a perturbation cannot be achieved by adding small random noise to $y_i$, the only possibility is to replace 0s by 1s and \emph{vice versa}. A perturb-all-data-at-once approach \citep{ElderIV2003} as for Gaussian data is thus not feasible. We explore ways to perturb Bernoulli-distributed $y$ below.

The equivalence of GDF and $k$ in the linear model encourages us to use GDF as a plug-in estimator of the number of parameters in the AIC computation.

\subsection{Cross-validation and a measure of model complexity} \label{sec:GDFCVequivalence}
% Alternatively: CV
Cross-validation is a standard approach to quantify predictive performance of a model, automatically accounting for model complexity \citep[by yielding lower fits for too simple and too complex models, e.g.][]{Wood2006,Hastie2009}. % and embraced particularly by machine-learning approaches and semi-parametric models.
Because each model is fitted repeatedly, cross-validation is computationally more expensive that the AIC, but the same problem arises for GDF: it requires many simulation to estimate it stably (see below). 

We decided to use the log-likelihood as measure of fit for the cross-validation \citep{Horne2006}, with the following reasoning. Let $f_{\theta}$ denote the model density of our data, \emph{y}, and let $f_t$ denote the true density. Thus the Kullback-Leibler (KL) divergence \citep{Kullback1951} is

\begin{equation}
\int \left(log f_t-log f_{\hat{\theta}}\right)f_t \,dy.
\end{equation}
AIC is an estimate of this values (in the process taking an expectation over the distribution of $\hat{\theta}$). 
Now $\int f_t log f_t\,dy$ is independent of the model, and the same for all models under consideration (and hence gets dropped from the AIC). So the important part of the KL-divergence is
\begin{equation}\label{eqn:KLdistshort}
-\int f_t log f_{\hat{\theta}} \,dy
\end{equation}
that is, the expected model log-likelihood, where the expectation is over the distribution of data not used in the estimation of $\hat{\theta}$. Expression~\ref{eqn:KLdistshort} can be estimated by cross-validation:
 \begin{equation}
\label{cvll}
\ell_{CV}=-\sum_{i = 1}^{K}log f_{\hat{\theta}^{\left[-i\right]}}\left(y_i\right),
\end{equation}
where $\ell_{CV}$ is the sum over \emph{K} folds of the log likelihood of the test subset $y^{\left[i\right]}$, given the trained model $f_{\hat{\theta}^{\left[-i\right]}}$, with parameter estimates $\theta^{\left[-i\right]}$. To put this on the same scale
as AIC we multiply by $-2$ to obtain the cross-validated deviance. % $\kappa$, which can be used in place of AIC for model selection and averaging.

With the assumption of AIC and (leave-one-out) cross-validation being asymptotically equivalent \citep{Stone1977} and given the definition of AIC, we argue that it should be possible to extract a measurement from $\ell_{CV}$ that quantifies model complexity. Hence,
\begin{equation}
\label{eqn:mcCV}
\begin{split}
\text{AIC}=-2\ell_m+2\hat{p} \approx -2\ell_{\text{CV}}\\
\hat{p} \approx \ell_m-\ell_{\text{CV}}.
\end{split}
\end{equation}
with $\hat{p}$ representing (estimated) model complexity, $\ell_m$ the maximum log-likelihood of the original (non-cross-validated) model and \emph{N} the number of data points. Embracing the small sample-size bias adjustment of AIC \citep{Sugiur1978, Hurvich1989} we get:
\begin{equation}
\label{eqn:mcCVc}
\begin{split}
\text{AICc}=-2\ell_m+2\hat{p}+\frac{2\hat{p}\left(\hat{p}+1\right)}{N-\hat{p}-1} \approx -2\ell_{\text{CV}}\\
\hat{p} \approx \frac{\left(\ell_m-\ell_{CV}\right)\left(N-1\right)}{\ell_{m}-\ell_{\text{CV}}+N}.
\end{split}
\end{equation}
Thus, we can compute both a cross-validation-based deviance that should be equivalent to the AIC, $-2\ell_{CV}$, as well as a cross-validation alternative to GDF, based on the degree of overfitting of the original model (eqns. \ref{eqn:mcCV} and \ref{eqn:mcCVc}).

This approach to computing model complexity is not completely unlike that of the DIC \citep{Spiegelhalter2002}, where $p_D$ represents the effective number of parameters in the model and is computed as the mean of the deviances in an MCMC-analysis minus the deviance of the mean estimates: $p_D = \overline{D(\theta)} - D(\bar{\theta}) = 2\ell(\bar{\theta}) -2\overline{\ell(\theta)}$ \citep{Wood2015}. In eqn.~\ref{eqn:mcCV} the likelihood estimate plays the role of $\ell(\bar{\theta})$, while the cross-validation estimates $\overline{\ell(\theta)}$.

\section{Implementing and evaluating the GDF-approach for normally and Bernoulli-distributed data}
We analysed simulated and real data sets using five different statistical models. Then we computed their GDF, disturbing $k$ data points at a time, with different intensity (only for normal data), and for different values of $k$.

\subsection{Data: simulated and real}
First, we evaluated the GDF-approach on normally distributed data following \citet{ElderIV2003},
%we chose three datasets with a binomially distributed response variable and one with a normally distributed response variable which can be regarded as the control run, since \citet{ElderIV2003} indicated functionality of GDF for normal regression problems.
deliberately using a relatively small data set: $N_{\text{norm}}=250$. The response $y$ was simulated as ($y \sim \mathcal{N}(\beta_0 + \beta_1x_1 + \beta_2x_1^2 + \beta_3x_2 + \beta_4x_3x_4, \sigma=1)$, with $\beta$-values of $-5, 5, -10,$ 10 and 10, respectively), with $x_{1 \ldots 4} \sim \mathcal{U}(0,1)$. (This simulation was repeated to control for possible influences of the data itself on the resulting GDF, but results were near-identical.) 

%-5 + 5*dats250$X1 - 10*(dats250$X1)^2 + 10*dats250$X2 + 10*dats250$X3*dats250$X4 + rnorm(250, mean=0, sd=2)

We simulated binary data with $N_{\text{binom}}=300$ (effective sample size ESS$_{\text{binom}}\approx 150$) and $y \sim \textrm{Bern}( \textrm{logit}^{-1}(\beta_0 + \beta_1x_1+ \beta_2x_1^2+ \beta_3x_2+ \beta_4x_3x4))$, (with $\beta$-values of $-6.66$, 5, $-10$, 10, respectively), with $x_{1 \ldots 4} \sim \mathcal{U}(0,1)$.

%rbinom(N300, size = 1, prob=plogis(-6.66 + 5*dats300$X1  - 10*(dats300$X1^2) + 10*dats300$X2 +10*dats300$X3*dats300$X4))

% Real-Life
Our real-world data comprised a fairly small data set ($N_{\emph{Physeter}}=261$, ESS$_{\emph{Physeter}}=115$), showing the occurrence of sperm whales (\emph{Physeter macrocephalus}) around Antarctica (collected during cetacean IDCR-DESS SOWER surveys), and a larger global occurrence data set ($N_{\emph{Vulpes}}=12722$, ESS$_{\emph{Vulpes}}=5401$) for the red fox (\emph{Vulpes vulpes}, provided by the International Union for Conservation of Nature (IUCN)). We pre-selected predictors as described in \citep{Dormann2010,Dormann2010a}, yielding the six and three (respectively) most important covariates. Since we are not interested in developing the best model, or comparing model performance, we did not strive to optimise the model settings for each data set.

\subsection{Implementing GDF for normally distributed data}
A robust computation of GDF is a little more problematic than eq.~\ref{gdf_vert} suggests. \citet{Ye1998} proposed to fit a linear regression to repeated perturbations of each data point, i.e. to $\left(\widehat{y'}_i-\widehat{y}\right)$ over $\left(y'_i-y_i\right)$ for each repeatedly perturbed data point $i$, calculating GDF as the sum of the slopes across all data points. As $y_i$ is constant for all models, and $\widehat{y}_i$ is constant for the non-stochastic algorithms (GLM, GAM), the linear regression simplifies to $\widehat{y'}_i$ over $y'_i$. For internally stochastic models we compute a mean $\overline{\widehat{y'}_i}$ as plugin for $\widehat{y}_i$, and therefore apply the procedure also to randomForest, ANN and BRT.
   
\citet{ElderIV2003} presents this so-called ``horizontal" method (the perturbed $y'$ and fitted $\widehat{y'}$ are stored as different columns in two separate matrices, which are then regressed row-wise) as more robust compared to the ``vertical" method (where for each perturbation a GDF is computed for each column in these matrices and later averaged across replicate perturbations). Convergence is poor when using the ``vertical" method, so we restricted the computation of GDF to the ``horizontal" method.

\subsection{GDF for binary data}
While normal data can be perturbed by adding normal noise (see next section), binary data cannot. The only option is to invert their value, i.e. $0\rightarrow1$ or $1\rightarrow0$. Clearly, we cannot perturb many data points simultaneously this way, as that will dilute any actual signal in the data. However, for large data sets it is computationally expensive to perturb all $n$ data points individually, and repeatedly (to yield values for the horizontal method, see above). We thus varied the number of data points to invert, $k$, to evaluate whether we can raise $k$ without biasing the GDF estimate.

%We further included a feed-forward Artificial Neural Network (ANN) computed from the \textbf{\textsf{nnet R}} package \citep{Venables2002}, parametrised using seven hidden units in a single layer and with weight decay chosen to be 0.1. In the case of gaussian data, where the ANN was used for regression, the outputs were switched to linear units (\textbf{\textsf{linout=TRUE}}).
%Lastly we fitted the data to a Boosted Regression Tree (BRT) model \citep{Friedman2001,Friedman2000} using the \textbf{\textsf{gbm R}} package \citep{Ridgeway2013}. Considerably differing from traditional regression based methods like GLMs or GAMs, we are dealing here with an ensemble method for which each individual model consists of a simple classification or regression tree.\\

\subsection{How many data points to perturb simultaneously?}
%Adapting GDF to binary data and their computation for machine-learning methods raises the question of how to best perturb the data. With the number of data points to perturb at a time, $k$, and in the Gaussian case the amplitude of the disturbance, there are two tuning parameters. 
With the number of points to perturb, $k$, the number of replicates for each GDF-computation ($n_\textrm{gdf}$) and the amplitude of disturbance (for the normal data), we have three tuning parameters when computing GDF.

First, we calculated GDF for the simulated data and the sperm whale dataset with $k$ taking up increasing values (from 1 to close to $n$,  or to ESS for binary data). Thus, random subsets of size $k$ of the response variable $y$ were perturbed, yielding $y'$. After each particular perturbation the model was re-fitted to $y'$. %We repeated this step $k+10$ times and saved the obtained $y'$ and $\widehat{y'}$ of all runs, forming the base of the final linear regression. 
To gain insight about the variance of the computed GDF values we repeated this calculation 100 times (due to very high computational effort, the number of reruns had to be limited to 10 for both randomForest and BRT).

The perturbation for the normally distributed $y$ were also drawn from a normal distribution $\mathcal{N}(0, 0.25\cdot\sigma_\textrm{simulation})$. We evaluated the sensitivity to this parameter by setting it to 0.125 and 0.5. 

%We then computed the GDF for the large real-life red fox data with \emph{k} considered most reasonable for binary data and the particular model.\\

\subsection{Modelling approaches}
We analysed the data using Generalised Linear Model (GLM), Generalised Additive Model (GAM), randomForest, feed-forward Artificial Neural Network (ANN) and Boosted Regression Trees (BRT). %See e.g. \citet{Guisan2002,Cutler2007,Lek1999,DeAth2007} for their relevance in ecology. 
%The \emph{a priori} selection was influenced by the individual model characteristics. The GLM as a fully deterministic, hence parametrised method allows for evaluation of the applied approaches. 
For the GLM, GDF should be identical to the trace of the Hessian (and the rank of the model), hence the GLM serves as benchmark. For GAMs, different ways to compute the degrees of freedom of the model have been proposed \citep{Wood2006}, while for the other three methods the intrinsic stochasticity of the algorithm (or in the case of ANN of the initial weights) may yield models with different GDFs each time.
%Described as ``nonparametric extension of GLMs" \citep[p.~587]{Yee1991}, but also referred to as ``semiparametric" methods \citep[e.g.][]{Guisan2006,Segurado2006}, generalised additive models (GAMs) \citep{Hastie1990,Wood2006} linked the deterministic with the blackbox approaches. 
%randomForest \citep{Breiman2001, Liaw2002} is advertised as an approach to find structure in high dimensional data. It is a so-called distribution-free method, as it makes no assumption about the distribution of predictor or response variables \citep{Cutler2007}.  

All models were fitted using R \citep{RCT2014} and packages gbm \citep[for BRTs: ][interaction depth=3, shrinkage=0.001, number of trees=3000, cv.folds=5]{Ridgeway2013}, mgcv \citep[for GAMs:][thin plate regression splines]{Wood2006}, nnet \citep[for ANNs:][size=7, decay=0.03 for simulated and decay=0.1 for real data, linout=TRUE for normal data]{Venables2002}, randomForest \citep{Liaw2002} and core package stats (for GLMs, using also quadratic terms and first-order interactions). R-code for all simulation as well as data are available on \url{https://github.com/biometry/GDF}.

\subsection{Computation of AIC and AIC-weights, from GDF and cross-validation}
In addition to the number of parameters, the AIC-formula requires the likelihood of the data, given the model. As machine-learning algorithms may minimise a score function different from the likelihood, the result probably differs from a maximum likelihood estimate. To calculate the AIC, however, we have to assume that the distance minimised by non-likelihood methods is proportional to the likelihood, otherwise the AIC would not be a valid approximation of the KL-distance. No such assumption has to be made for cross-validation, as $\ell_\text{CV}$ here only serves as a measure of model performance. For the normal data, we compute the standard deviation of the model's residuals as plug-in estimate of $\sigma$.

For the binary data, we use the model fits as probability in the Bernoulli distribution to compute the likelihood of that model.
We then calculated the AIC for all considered models based on their GDF-value. Due to the small sample sizes, we used AICc \citep{Sugiur1978, Hurvich1989}:
\begin{equation}
\label{aicc}
\textrm{AICc}=-2\ell_m+2\text{GDF} + \frac{\text{GDF}\left(\text{GDF}+1\right)}{N-\text{GDF}-1}.
\end{equation}
%with $n$ denoting the number of data points. % Its use for large datasets is unproblematic because $\lim\limits_{n \ll \text{GDF}}{\frac{\text{GDF}\left(\text{GDF}+1\right)}{n-\text{GDF}-1}}=0$.

We used (10-fold) cross-validation, maintaining prevalence in the case of binary data, to compute the log-likelihood of the cross-validation, yielding $\ell_\textrm{CV}$. To directly compare it to AICc, we multiplied $\ell_\textrm{CV}$ with $-2$. The cross-validation automatically penalises for overfitting by making poorer predictions. 

For model averaging, we computed model weights  $w_m$  for each model $m$, once for the GDF-based AICc and for the cross-validation log-likelihood, using the equation for Akaike-weights \citep[p.~75]{Turkheimer2003,Burnham2002}: 
\begin{equation}
\label{AICw}
w^{\text{AICc}}_m=\frac{e^{-\frac{1}{2}\Delta_m^{\text{AICc}}}}{\sum_{r=1}^{n}e^{-\frac{1}{2}\Delta_r^{\text{AICc}}}},
\end{equation}   
where $\Delta_m^{\text{AICc}}=\text{AICc}_m - \text{AICc}_{\min}$, taking the smallest AICc, i.e. the AICc of the best of the candidate models as $\text{AICc}_{\min}$; \emph{n} is the number of models to be averaged over. 

The same idea can be applied to cross-validated log-likelihoods, so that
\begin{equation}
\label{CVw}
w^{\text{CV}}_m=\frac{e^{\Delta_m^{\text{CV}}}}{\sum_{r=1}^{n}e^{\Delta_r^{\text{CV}}}},
\end{equation}
where $\Delta_m^{\text{CV}}=\ell_{\text{CV}_{\max}} - \ell_{\text{CV}_m}$, with $\ell_{\text{CV}_{\max}}$ being the largest cross-validated log-likelihood in the model set.

%\textcolor{red}{Severin: diese Formel war falsch! Ich hoffe, Du hast sie richtig implementiert?!} % yep!

%Finally, we visually compare the AICc based on the computed GDF with the cross-validated deviance $\kappa$ and check the equivalence of model weights and model complexity retrieved by the both approaches. 

% ------------------------------------------------------------ %
\section{Results}

\subsection{GDF configuration analysis}
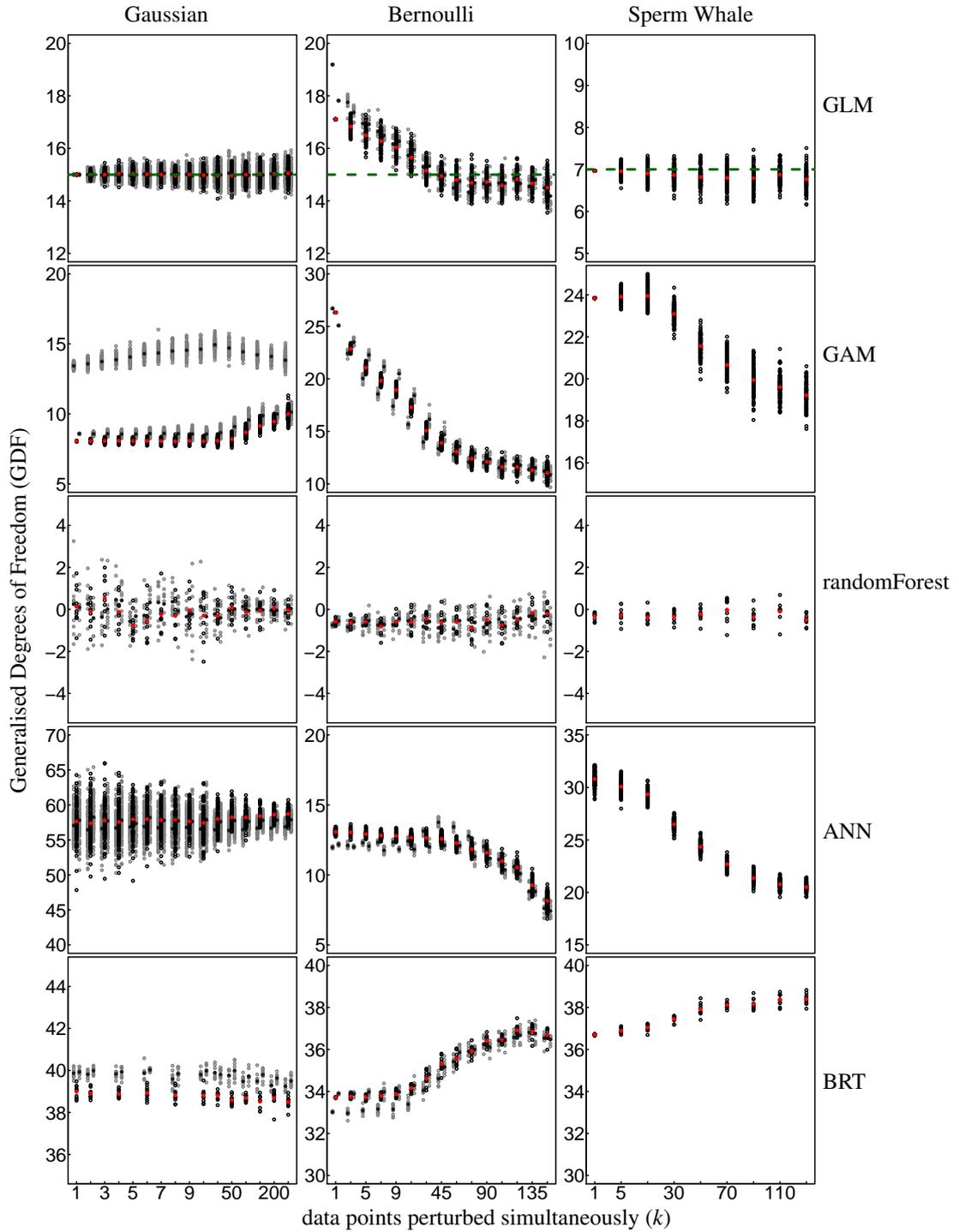
\begin{figure}
\centering
\includegraphics[keepaspectratio, width=.82\textwidth] {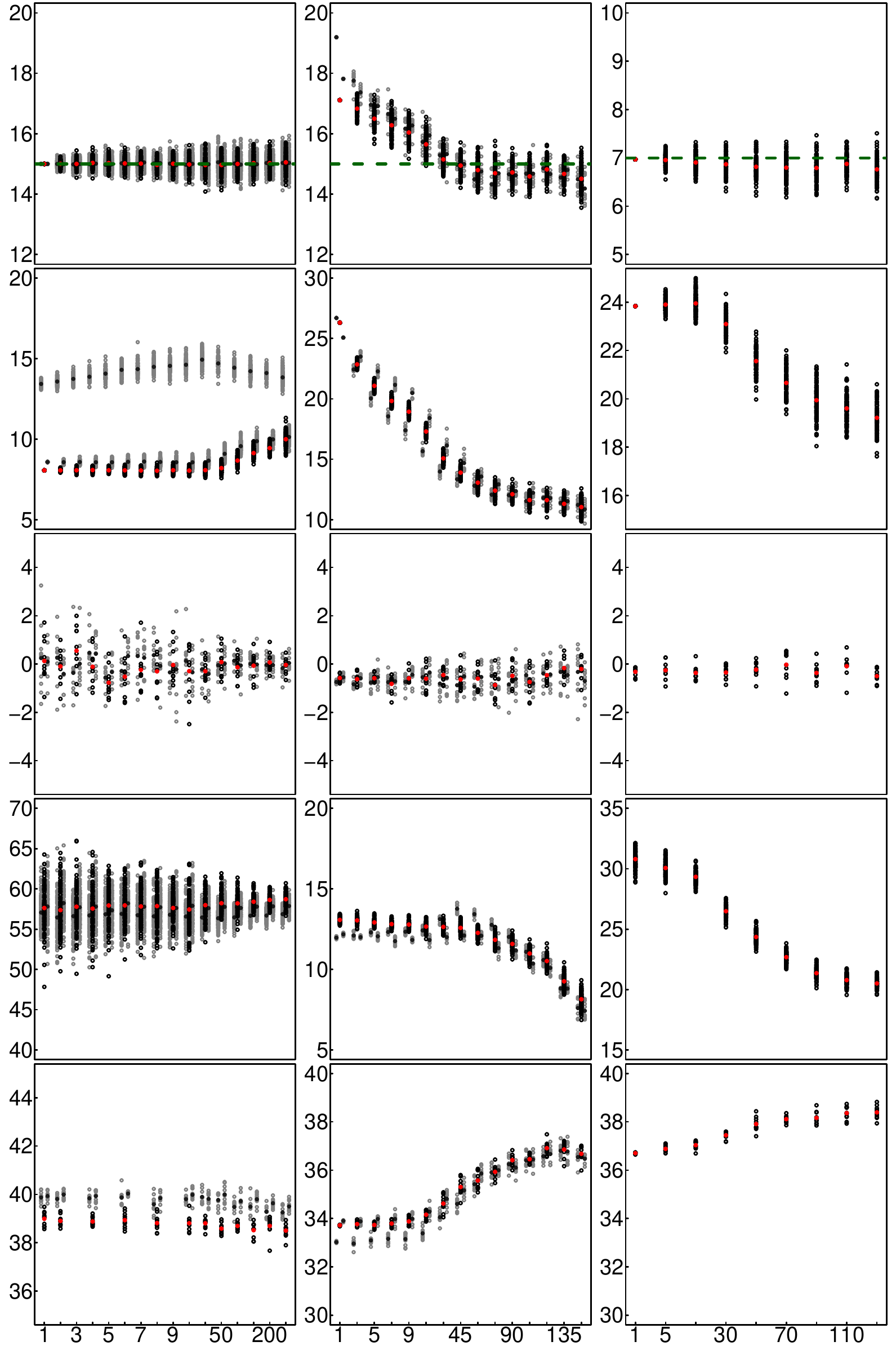}
\put(-305,520){Gaussian}
\put(-190,520){Bernoulli}
\put(-85,520){Sperm Whale}
\put(-0,480){GLM}
\put(-0,370){GAM}
\put(-0,270){randomForest}
\put(-0,160){ANN}
\put(-0,50){BRT}
\put(-230,-10){ data points perturbed simultaneously (\emph{k})}
\put(-355,180){\rotatebox{90}{Generalised Degrees of Freedom (GDF)}}
\caption{Models' GDFs as a function of the numbers of  data points perturbed (\emph{k}) (ranked abscissa) for the simulated Gaussian (left), Bernoulli (centre) and sperm whale data (right column) and the five model types (rows). The filled red \tikzcircle[red, fill=red]{1.5pt} and black dot \tikzcircle[black, fill=black]{1.5pt} represent the mean GDF for the two replications at each level of \emph{k} (in light-grey open dots \tikzcircle[grey, fill=white]{1.5pt} and black open dots \tikzcircle[black, fill=white]{1.5pt}). The dashed line in the first row \textcolor{darkgreen}{\xdash[0.5em]}~\textcolor{darkgreen}{\xdash[0.5em]}~\textcolor{darkgreen}{\xdash[0.5em]} represents the number of model parameters of the GLM. %For the GAM (second row), \textcolor{darkgreen}{\xdash[0.1em]}~\textcolor{darkgreen}{\xdash[0.1em]}~\textcolor{darkgreen}{\xdash[0.1em]}~\textcolor{darkgreen}{\xdash[0.1em]} and \textcolor{darkgreen}{\xdash[0.5em]}~\textcolor{darkgreen}{\xdash[0.1em]}~\textcolor{darkgreen}{\xdash[0.5em]}~\textcolor{darkgreen}{\xdash[0.1em]} are the edfs of the two simulation.
}
\label{gdf_config}
\end{figure}

For normally distributed data, increasing the number of points perturbed simultaneously typically slightly increased the variance of the Generalised Degrees of Freedom calculated for the model (Fig.~\ref{gdf_config} left column, GLM, GAM, but not for randomForest and ANN). For GLM and GAM, GDF computations yielded exactly the same value as the model's rank (indicated by the dashed horizontal line).

For simulated Bernoulli data (Fig.~\ref{gdf_config} central column), we also observed an effect of the number of points perturbed on the actual GDF value, which decreased for GLM, GAM and ANN, but increased for BRTs with the number of data points perturbed. Several data points needed to be perturbed ($> 20$) to yield an approximately correct estimate. More worryingly, GDF depended non-linearly on the number of data points perturbed, with values varying by a factor of 2 for GAM and ANN. For GAM the sensitivity occurs because the smoothing parameter selection is sensitive to the quite severe information loss as more and more data are perturbed. GLM and BRT yielded more consistent but still systematically varying GDF-estimates.

The same pattern was observed for the sperm whale data (Fig.~\ref{gdf_config}, right column). For the GLM, there was still some bias observable, also in the sperm whale-case study's GLM. We attribute it to the fact that by perturbing the binary data we also alter the prevalence.

The two replicate simulations yielded consistent estimates, except in the case of the normal GAM and normal BRT. This suggests that both methods ``fitted into the noise'' specific to the data set, while the other methods did not. We did not observe this phenomenon with the binary data.

For randomForest, GDF-estimates centre around $0$, meaning that perturbations of the data did not affect the model predictions. This is to some extent explicable by the stochastic nature of randomForest, giving different predictions when fitted to the same data. We interpret the value of $0$ as the perturbations creating less variability than the intrinsic stochasticity of this approach. 

This is not a general feature of stochastic approaches, as in neural networks and boosted regression trees the intrinsic stochasticity seems to be much less influential, and both approaches yielded relatively consistent GDF-values. The actual GDF estimate of course depends on the settings of the methods and should not be interpreted as representative.

\begin{figure}
\centering
\includegraphics[keepaspectratio, width=.82\textwidth] {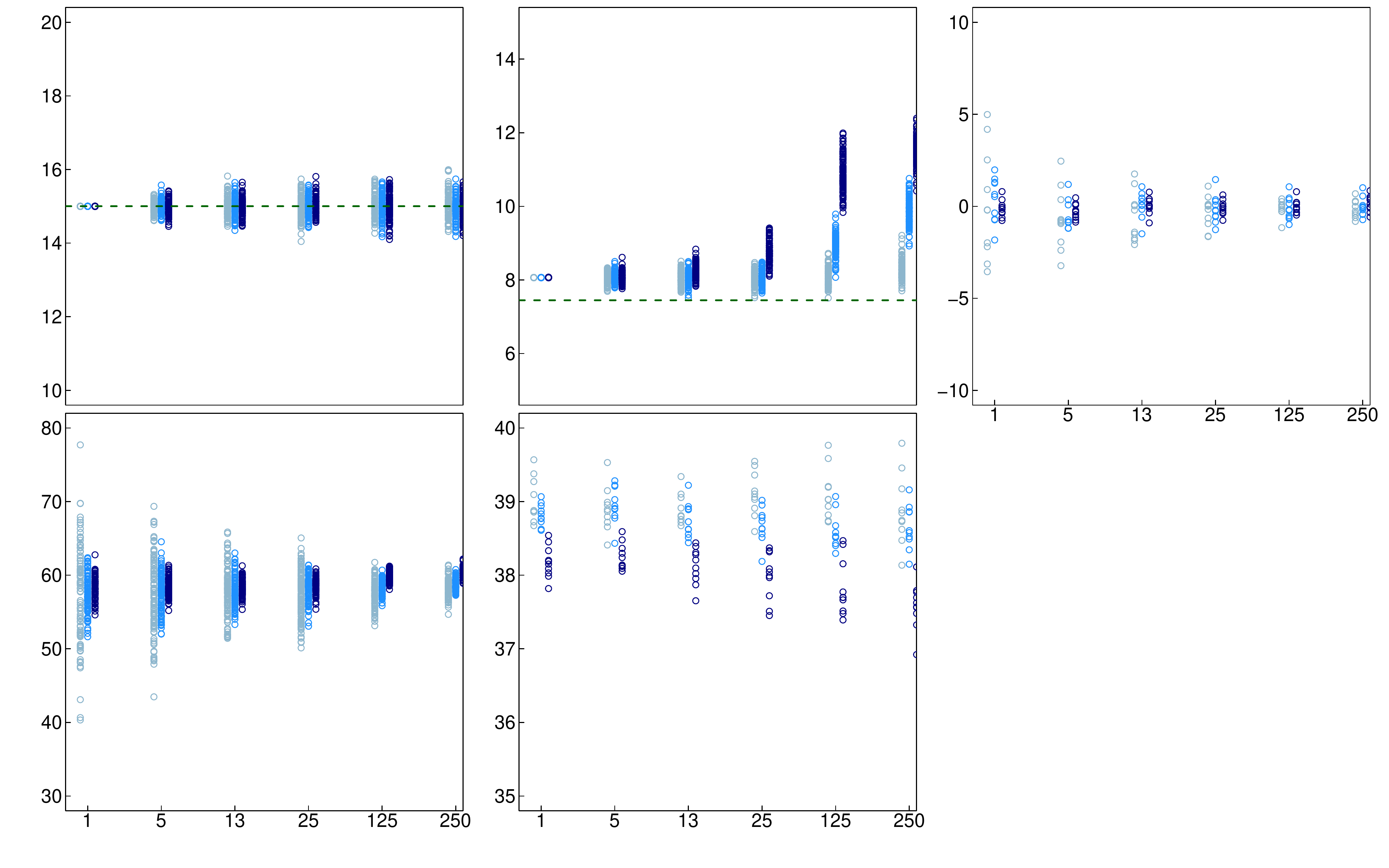}
\put(-325,200){GLM}
\put(-213,200){GAM}
\put(-100,200){randomForest}
\put(-325,100){ANN}
\put(-213,100){BRT}
\put(-230,-10){perturbed data points at once (\emph{k})}
\put(-350,30){\rotatebox{90}{Generalised Degrees of Freedom (GDF)}}
\caption{\textbf{Perturbation magnitude:} Estimated GDF along an increasing numbers of perturbed data points $(k)$ for the simulated Gaussian data. For each $k$ there are 100 GDF replications computed with $\sigma=0.125\sigma_{y}$ (\tikzcircle[lightblue, fill=lightblue]{1.5pt}), $\sigma=0.25\sigma_{y}$ (\tikzcircle[blue, fill=blue]{1.5pt}) and $\sigma=0.5\sigma_{y}$ (\tikzcircle[darkblue, fill=darkblue]{1.5pt}), hence lower to higher perturbation magnitude. The dotted green line \textcolor{darkgreen}{\xdash[0.5em]}~\textcolor{darkgreen}{\xdash[0.5em]}~\textcolor{darkgreen}{\xdash[0.5em]}~\textcolor{darkgreen}{\xdash[0.5em]} represents the model's rank (GLM) and the sum of estimated degrees of freedom (GAM). Note that x-axis is rank-transformed. 
\label{gdf_config_noise}}
\end{figure}

To compute the GDF for normally distribute data, we have to choose the strength of perturbation. The GDF-value is robust to this choice, unless many data points are disturbed (Fig.~\ref{gdf_config_noise}). Only for BRT does increasing the strength of perturbation lead to a consistent, but small, decrease in GDF-estimates, suggesting again that BRTs fit into the noise.

%The impact of the perturbation magnitude in the case of the normally distributed data is shown in Fig.~\ref{gdf_config_noise}. We do this in combination with increasing values of \emph{k}. For each \emph{k} from left to right and from light blue to dark blue dots $\sigma$ in $\operatorname{N}\left(\mu=0,\sigma\right)$, hence the perturbation magnitude is increasing by a factor of two.
%For the GLM, neither a decrease nor an increase of the added random noise to the data seems to have an effect on the resulting GDF values. These observations do not apply unconditionally to the other considered models. Especially in the case of the GAM we note quite a difference for varying noise intensities considering large \emph{k}. It seems like the in fig.~\ref{gdf_config} described drift of GDF for larger \emph{k} can be antagonised by decreasing $\sigma$.
Estimates of GDF for randomForest and ANN (but not BRT) respond with decreasing variance to increasing the strength of perturbation. %Showing a randomly, slightly varying prediction with each run, these methods feature an internal random noise. 
Increasing the intensity of perturbation seems to overrule their internal stochasticity. %For BRT, we observe a rather stable development for lower perturbation values and a drift for higher \emph{k} and higher $\sigma$.

% How many replications are needed?
It seems clear from the results presented so far that no single best perturbation strength and optimal proportion of data to perturb exists for all methods. For the following analyses, we use, for normal data, perturbation $= 0.25\sigma_y$ and $k = N$ (except for GAM, where $k=0.2N$); and for Bernoulli data $k=0.5N$ (except for BRT and ANN, where $k=0.04N$).
% Inferrence from first part of analysis
%On the basis of this first part of the analysis, we infer that $\sigma=0.25\sigma_{y_{norm}}$ and $k=250=N$ (except for GAM, where $k=50=0.2N$) for the simulated gaussian data, $k=150=0.5N$ for GLM, GAM and randomForest , for ANN $k=30=0.1N$ and for BRT $k=15=0.05N$ for the simulated binomial data and $k=130=0.5N$ for GLM, GAM and randomForest and $k=10=0.04N$ for ANN and BRT and the sperm whale data produce the most stable results and are therefore appropriate to compute the GDF. The configuration of both binary datasets form the basis for the large red fox data GDF computation.

\begin{figure}
\centering
\includegraphics[keepaspectratio, width=0.34\textwidth] {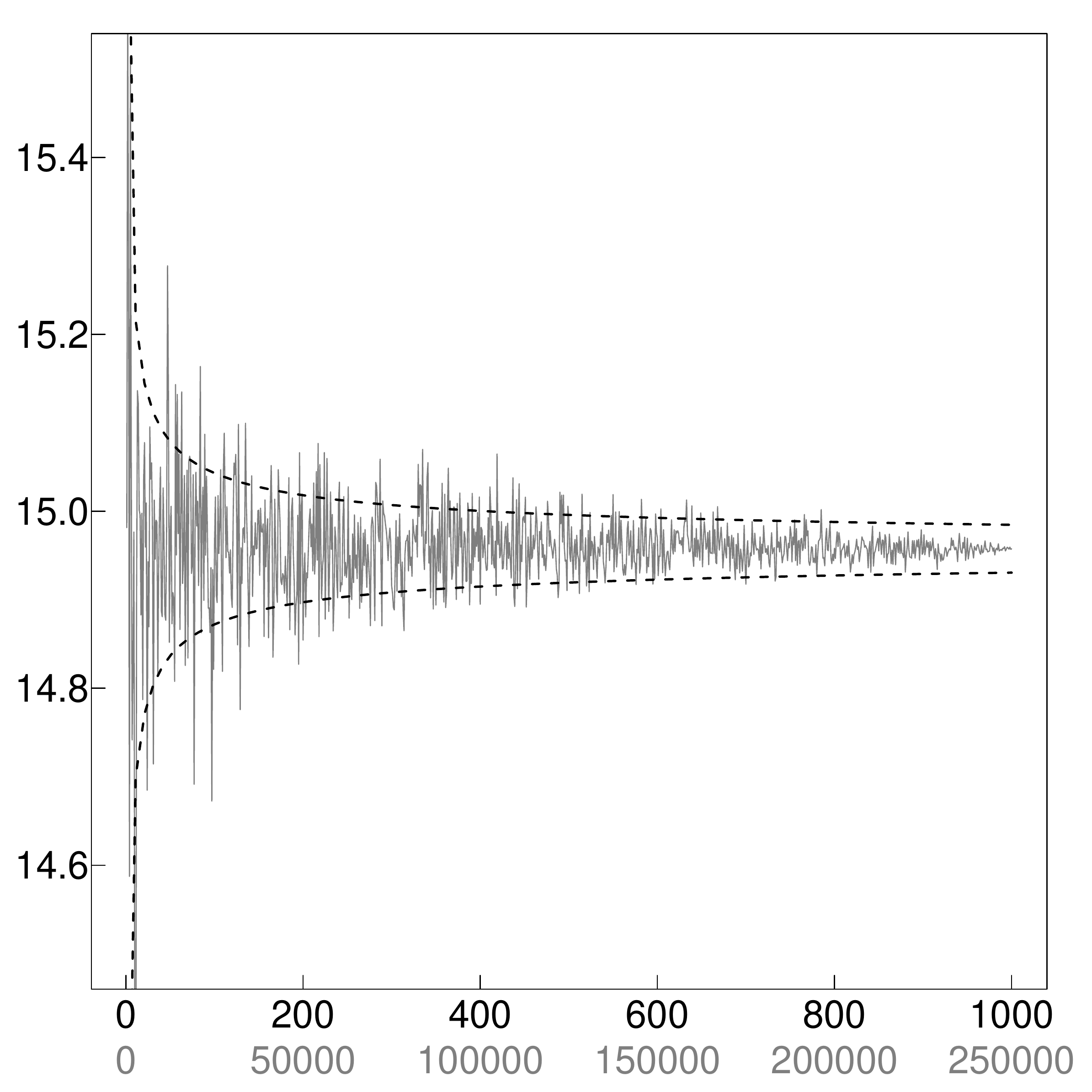}\hspace{0.3cm}
\includegraphics[keepaspectratio, width=0.34\textwidth]{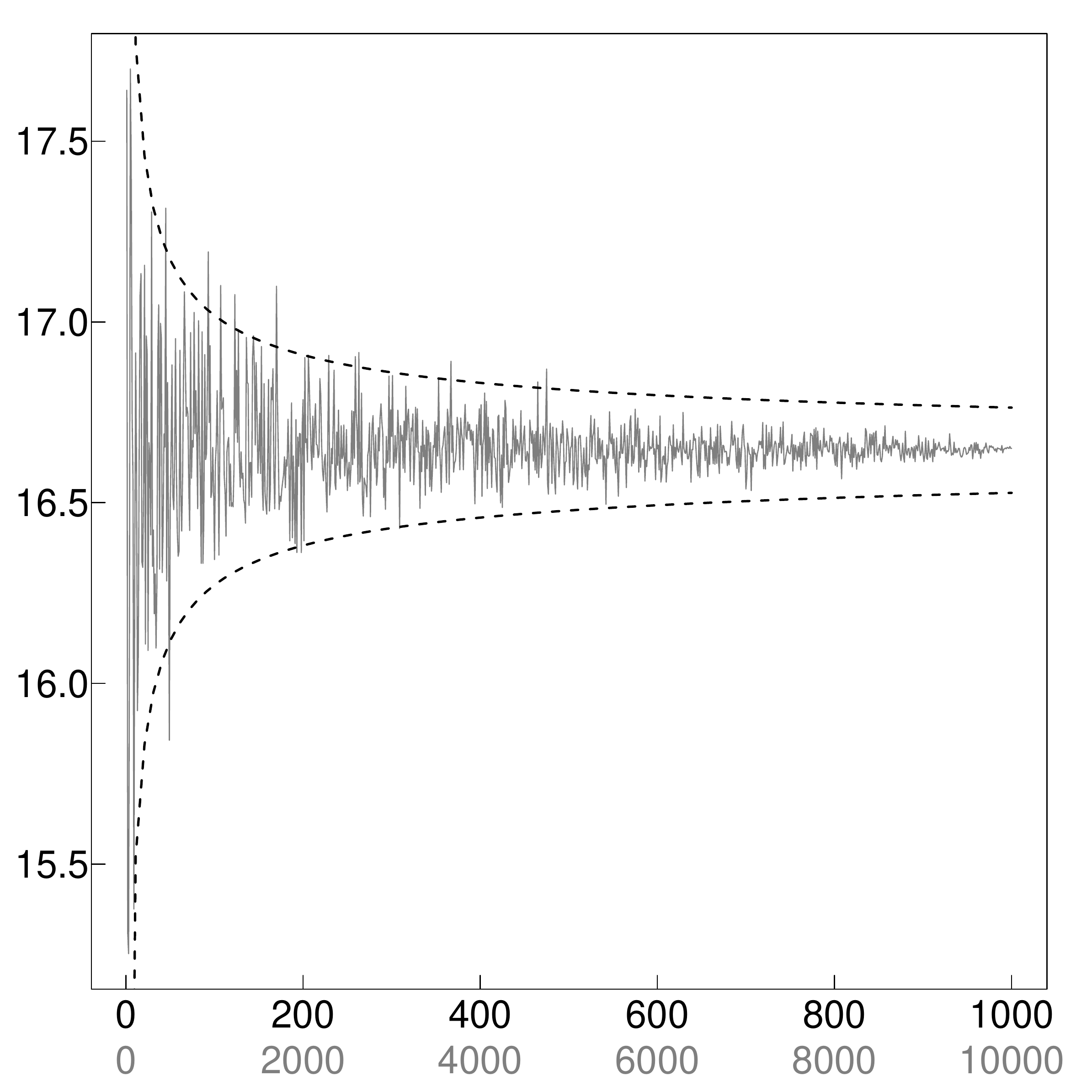}
\put(-210,-15){Number of Replicates/ \textcolor{gray}{Model runs}}
\put(-305,65){\rotatebox{90}{GDF}}
\put(-148,65){\rotatebox{90}{$\Delta \ell^\text{CV}_m$}}
\caption{Development of mean Generalised Degrees of Freedom (GDF, left) and cross-validation log-likelihood-differences ($\Delta \ell^\text{CV}_m$, right) over 1000 replicates (250000 and 10000 model runs, respectively). GDF computation with $k=50$ data points perturbed per model run and 50 internal replicates for the Gaussian simulation data ($N=250$), i.e. 250 model evaluations per replicate. $\Delta \ell^\text{CV}_m$ derives from 10-fold cross-validation, i.e. 10 model evaluations per replicate. The dashed lines display the mean GDF  and $\Delta \ell^\text{CV}_m$, respectively, of all replicates $\pm$ one standard error. \label{replicates}}
\end{figure}

\subsection{Efficiency of GDF and cross-validation computations}
Both GDF and cross-validation require multiple analyses. For the GDF, we need to run several perturbations, and possibly replicate this many times. For cross-validation, we may also want to repeatedly perform the 10-fold cross-validation itself to yield stable estimates. The mean GDF and CV-log-likelihood ($\ell_{CV}$) over 1-1000 replicates are depicted in Fig.~\ref{replicates}. With 100 runs, both estimates have stabilised, but 100 runs for GDF represent 25000 model evaluations (due to the $k=50$ perturbations and 50 internal replicates for a data set of $N=250$), while for 10-fold cross-validation these represent only 1000 model evaluations, making it 25 times less costly.

%Furthermore, the left graphic in fig.~\ref{replicates} suggests one uses the mean over at least 100 GDF replicates in order to increase their stability. Similar applies to cross-validation (right graphic in fig.~\ref{replicates}). Likewise, the standard error seems to be substantially decreased after 100 replicates, even though the actual variance is far higher because of the absolute value of $\ell_{\text{CV}}$. 
%However, the same amount of GDF replicates involve incomparably more costs. Here the computation of only one replication requires 250 model runs, assumed $k=50$, $pass=50$ and $N=250$, whereas the 10-fold cross-validation needs 10 re-runs for one replication.\\

\subsection{GDF-based AIC vs Cross-Validation}
We analysed four data sets with the above settings for GDF and cross-validation, two simulated (Gaussian and Bernoulli) and two real-world data sets (sperm whale and red fox geographic distributions). % \textcolor{red}{The settings for the simulations were ....}\textcolor{green}{What do you mean here? Just copying from above?}

%As was mentioned before, there may be a way to derive a measure of model complexity from the cross-validated log-likelihood (MC$_\text{CV}$) (see~eq.~\ref{mcCV} and eq.~\ref{mcCVc}). 
%Table \ref{mcomplex} however, illustrates that the comparison of GDF and MC$_\text{CV}$ is rather not satisfactory. Whereas the dimensional ranking often coincides for both GDF and MC$_\text{CV}$, especially in the case of gaussian simulation data, the actual values differ substantially. Remarkable are once again the results concerning the randomForests. Except for the large red fox data even MC$_\text{CV}$ exhibits values close to zero, sometimes even below.     

Both Generalised Degrees of Freedom (GDF$_m$) and cross-validation log-likelihood-differences ($\Delta \ell^\text{CV}_m$) measure model complexity in an asymptotically equivalent way (section \ref{sec:GDFCVequivalence}). For finite data sets both approaches yield identical rankings of model complexity but are rather different in absolute value (eqn.~\ref{mcomplex}). Particularly the red fox-case study yields very low GDF-estimates for GLM, GAM and randomForest, while their cross-validation model complexity is much higher. 

%Based on the GDF using \emph{k} (and in the Gaussian case $\sigma$), which we suppose to be most accurate, AIC (or rather AICc) can be calculated for all datasets, including the large red fox data. We compare the obtained values with the repeated 10-fold cross-validated deviance $\kappa$ by visually analysing their relationship. Being on the same scale the corresponding values ideally occur closely to the bisector of the first quadrant indicating identity of $x=\text{AICc}_m$ and $y=\kappa_m$.

% model complexity
\begin{table}
\centering
\caption{Measures of model complexity (mean $\pm$ standard deviation): Generalised Degrees of Freedom (GDF) and $\Delta \ell^\text{CV}_m$ derived from cross-validation (see section~\ref{sec:GDFCVequivalence}) for Gaussian and Bernoulli simulation data and real-world sperm whale and red fox distribution data. The true ranks for the GLMs are 15, 15, 7 and 8, respectively.}
\label{mcomplex}
\begin{tabular}{ccccc} 
\toprule
Model & GDF & $\Delta \ell^\text{CV}_m$ &  GDF &  $\Delta \ell^\text{CV}_m$\\  %table heading
\midrule
& \multicolumn{2}{c}{\textsc{Gaussian simulation}} & \multicolumn{2}{c}{\textsc{Bernoulli simulation}} \\
GLM & $15.1\pm 0.30$& $16.7\pm 1.97$& $14.5\pm 0.41$& $19.4\pm 3.18$\\
GAM & $10.0\pm 0.41$& $10.7\pm 2.09$& $11.0\pm 0.58$& $27.0\pm 3.82$\\ 
randomForest & $-0.04\pm 0.33$& $2.8\pm 2.67$& $-0.21\pm 0.41$& $1.3\pm 2.10$\\
ANN & $58.7\pm 0.82$& $69.4\pm 5.76$& $12.6\pm 0.25$& $12.1\pm 1.67$\\
BRT & $38.5\pm 0.28$& $45.1\pm 2.17$& $34.2\pm 0.17$& $20.3\pm 1.29$\\ 
%\midrule
& \multicolumn{2}{c}{\textsc{Sperm whale}} & \multicolumn{2}{c}{\textsc{Red fox}} \\
%\midrule
GLM & $6.8\pm 0.24$& $7.2\pm 1.40$& $7.9\pm 0.38$& $37.39\pm 13.0$\\
GAM & $19.2\pm 0.59$& $29.8\pm 7.63$& $9.0\pm 0.21$& $37.08\pm 10.4$\\ 
randomForest  & $-0.50\pm 0.26$& $-1.1\pm 2.66$& $1.11\pm 5.83$& $31.7\pm 13.22$\\
ANN & $29.3\pm 0.49$& $35.0\pm 4.96$& $50.3\pm 0.66$& $64.2\pm 80.69$\\
BRT & $37.0\pm 0.17$& $26.0\pm 1.29$& $59.0\pm 0.12$& $54.6\pm 2.30$\\ 
%%\midrule              
%GLM & $15.05\pm 0.30$& $16.65\pm 1.97$& $14.51\pm 0.41$& $19.37\pm 3.18$\\
%GAM & $9.99\pm 0.41$& $10.72\pm 2.09$& $11.04\pm 0.58$& $27.02\pm 3.82$\\ 
%randomForest & $-0.04\pm 0.33$& $2.83\pm 2.67$& $-0.21\pm 0.41$& $1.29\pm 2.10$\\
%ANN & $58.72\pm 0.82$& $69.39\pm 5.76$& $12.64\pm 0.25$& $12.08\pm 1.67$\\
%BRT & $38.50\pm 0.28$& $45.06\pm 2.17$& $34.16\pm 0.17$& $20.29\pm 1.29$\\ 
%%\midrule
%& \multicolumn{2}{c}{\textsc{Sperm whale}} & \multicolumn{2}{c}{\textsc{Red fox}} \\
%%\midrule
%GLM & $6.76\pm 0.24$& $7.21\pm 1.40$& $7.92\pm 0.38$& $37.39\pm 13.00$\\
%GAM & $19.22\pm 0.59$& $29.82\pm 7.63$& $8.97\pm 0.21$& $37.08\pm 10.41$\\ 
%randomForest  & $-0.50\pm 0.26$& $-1.11\pm 2.66$& $1.11\pm 5.83$& $31.73\pm 13.22$\\
%ANN & $29.33\pm 0.49$& $34.96\pm 4.96$& $50.26\pm 0.66$& $64.24\pm 80.69$\\
%BRT & $37.04\pm 0.17$& $25.95\pm 1.29$& $59.00\pm 0.12$& $54.57\pm 2.30$\\ 
\bottomrule
\end{tabular}
\end{table}

Model complexity is only one term in the AIC-formula (eqn.~\ref{eqn:mcCV}) and for large data sets the log-likelihood will dominate. To put the differences between the two measures of model complexity into perspective, we computed AIC$_\text{GDF}$ for all data sets and compared it to the equivalent cross-validation deviance ($-2 \ell_\text{CV}$).

\begin{figure}
\centering
\includegraphics[keepaspectratio, width=0.41\textwidth] {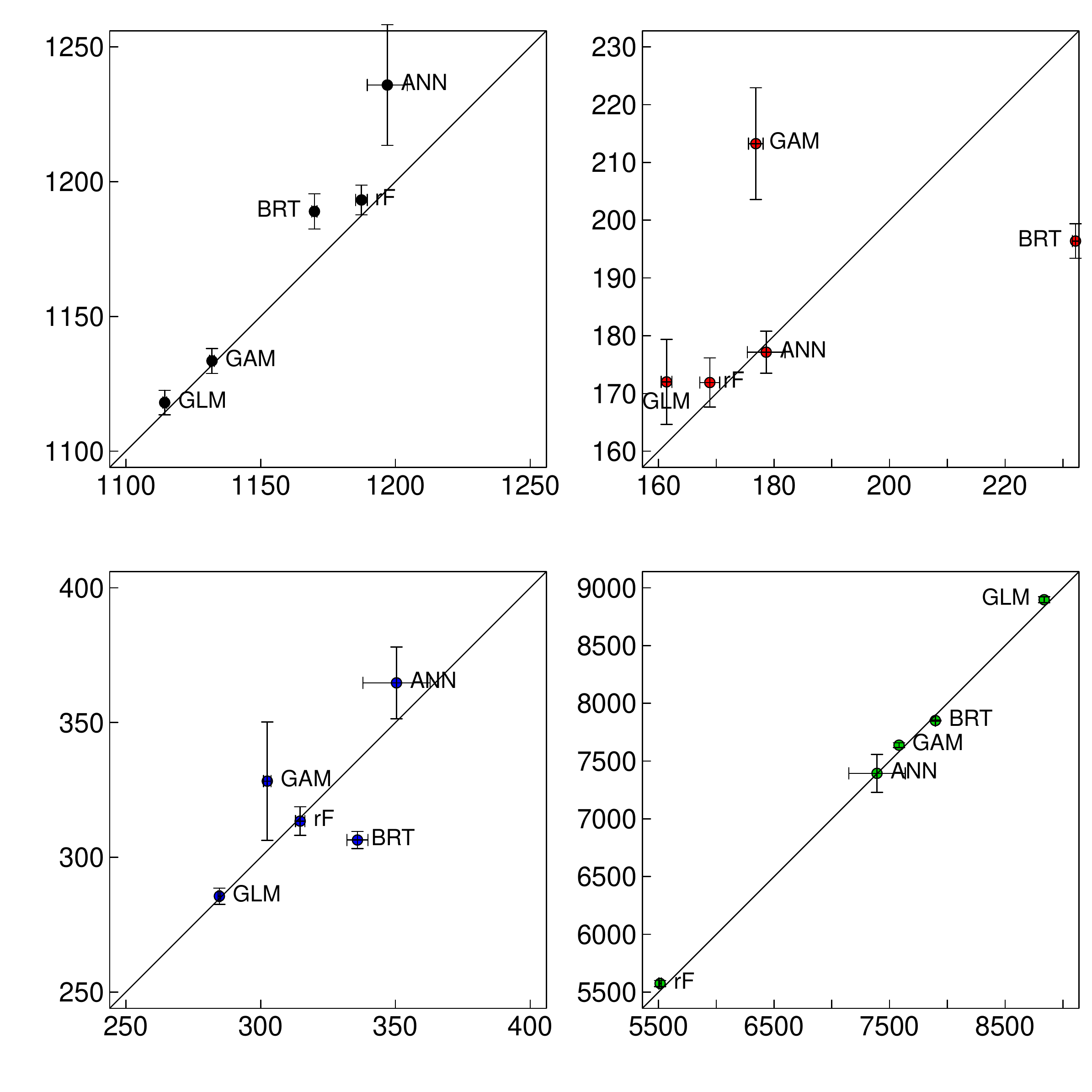}
\includegraphics[keepaspectratio, width=0.41\textwidth] {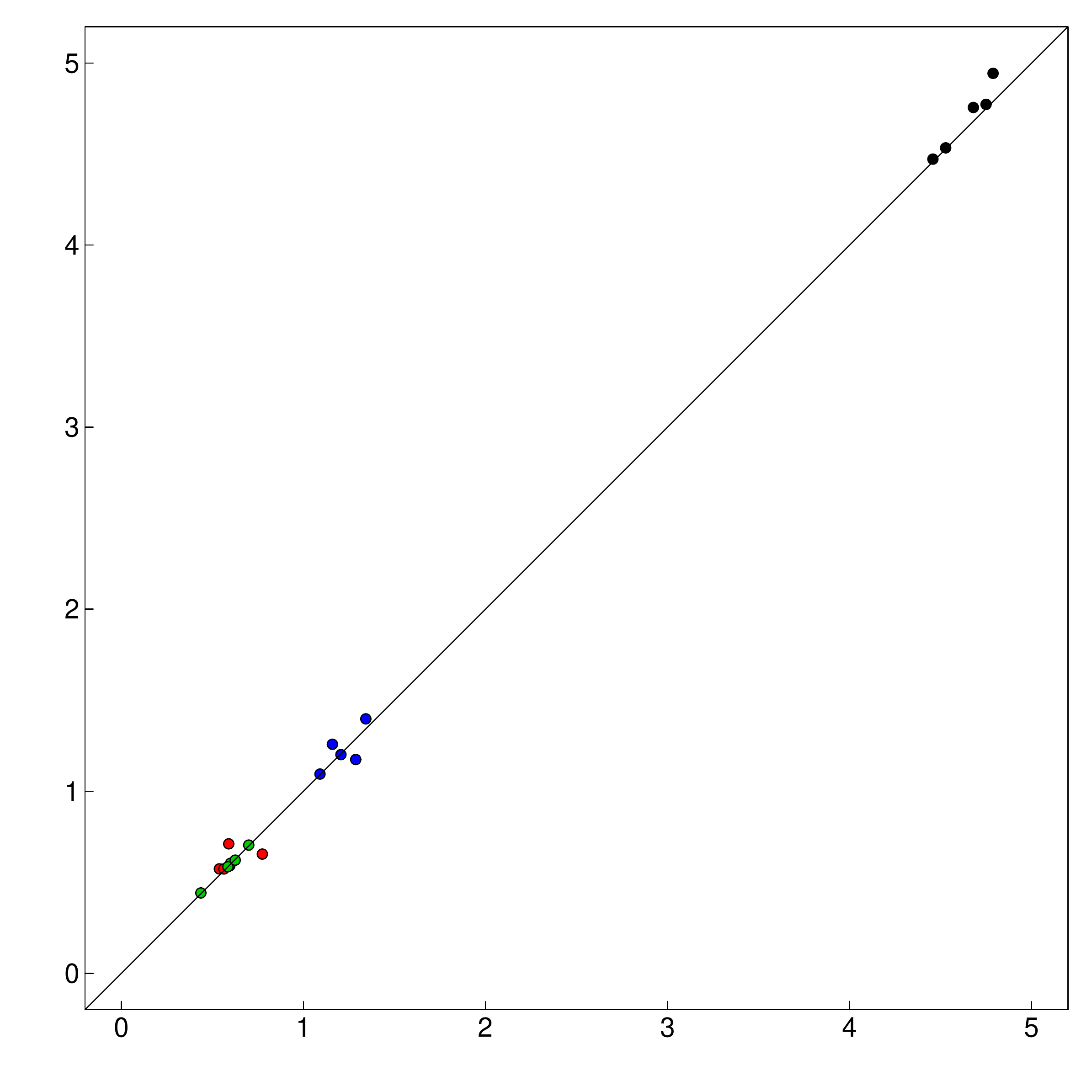}
\put(-195,-5){GDF-based AIC}
\put(-350,22){\rotatebox{90}{Cross-validated deviance $-2 \ell^\text{CV}$}}
\caption{Cross-validated deviance versus GDF-based AICc. Error bars display $\pm$ one standard error (sometimes too small to be visible). The identity line %\xdashh[1.5em] 
($y=x$) indicates the equivalence of both measures. Colours represent the four datasets: Gaussian simulated (top-left, \tikzcircle[black, fill=black]{1.5pt}) and Bernoulli simulated (top-centre, \tikzcircle[black, fill=red]{1.5pt});  sperm whale (bottom-left, \tikzcircle[black, fill=blue]{1.5pt}) and red fox (bottom-centre, \tikzcircle[black, fill=green3]{1.5pt}) data. In the right panel per datum GDF and $-2 \ell^\text{CV}$, respectively, i.e. divided by the number of data points.}\label{fig:aic_cv}
\end{figure}

Across the entire range of data sets analysed both approaches yield extremely similar results (Fig.~\ref{fig:aic_cv} right). Within each data set, however, the pattern is more idiosyncratic, revealing a high sensitivity to low sample size ($N<1000$, i.e. all data sets except the red fox).

\subsection{GDF, $\Delta \ell^{\text{CV}}_m$ and model weights}
For all four datasets one modelling approach always substantially outperforms the others, making model averaging an academic exercise. We compare model weights (according to eqns~\ref{AICw} and \ref{CVw}) purely to quantify the differences in AIC$_\text{GDF}$ and cross-validation deviance for model averaging. Only for the Bernoulli simulation is the difference noticeable (Table~\ref{weights}). Here GLM and randomForest share the model weight when quantified based on cross-validation, while for the GDF-approach GLM takes all the weight.  %Nevertheless, the obtained AIC and cross-validated deviance can be transformed to Akaike weights ($w_\text{AIC}$) and cross-validation weights ($w_\text{CV}$), respectively. We find the results of both approaches contrasted with each other in Table~\ref{weights}. 

%The weighting coefficients take up values between zero and one. A higher value indicates a relatively better model, which has more impact on the averaged global predictions. In the case of the red fox data only the randomForest constitutes the averaged model, since the weights ($w_\text{AIC}$ and $w_\text{CV}$) of the four other candidates are negligibly small. Concerning the other datasets, the averaged model consists almost solely of the GLM, except for the simulated Bernoulli data, where $w_\text{CV}$ of both the randomForest and GLM are close to 0.5. 
%This, however, is not the focus of interest here. Table~\ref{weights} rather illustrates the possible outcome of conventional Akaike weights and cross-validation weights.  

\begin{table}
\centering
\caption{\textbf{Model weights}: Comparison of Akaike weights ${w_\text{AIC}}$ and cross-validation weights ${w_\text{CV}}$ (see eq.~\ref{CVw}) for Gaussian and Bernoulli simulation data and real-life sperm whale and red fox abundance data. The ``surprisingly'' good performance of GLMs in the sperm whale case study remains unexplained, but is fully reproducible. \label{weights}}
\begin{tabular}{ccccc} 
\toprule
\textsc{\textbf{Model}} & \bfseries ${w_\text{AIC}}$ & \bfseries ${w_\text{CV}}$ & \bfseries ${w_\text{AIC}}$ & \bfseries ${w_\text{CV}}$\\  %table heading
\midrule
& \multicolumn{2}{c}{\textsc{Gaussian simulation}} & \multicolumn{2}{c}{\textsc{Bernoulli simulation}} \\
%\midrule
GLM & $0.9998$& $0.9996$& $0.9762$& $0.4682$\\
GAM & $1.548\cdot10^{-4}$& $4.390\cdot10^{-4}$& $4.310\cdot10^{-4}$& $5.185\cdot10^{-10}$\\ 
rF  & $1.407\cdot10^{-16}$& $4.772\cdot10^{-17}$& $2.319\cdot10^{-2}$& $0.4960$\\
ANN & $1.158\cdot10^{-18}$& $2.632\cdot10^{-26}$& $1.759\cdot10^{-4}$& $3.578\cdot10^{-2}$\\
BRT & $8.647\cdot10^{-13}$& $4.001\cdot10^{-16}$& $4.131\cdot10^{-16}$& $2.381\cdot10^{-6}$\\ 
%\midrule
& \multicolumn{2}{c}{\textsc{Sperm whale}} & \multicolumn{2}{c}{\textsc{Red fox}} \\
%\midrule
GLM & $0.9999$& $0.9997$& $0$& $0$\\
GAM & $1.373\cdot10^{-4}$& $5.387\cdot10^{-10}$& $0$& $0$\\ 
rF  & $3.079\cdot10^{-7}$& $8.966\cdot10^{-7}$& $1$& $1$\\
ANN & $5.369\cdot10^{-15}$& $6.558\cdot10^{-18}$& $0$& $0$\\
BRT & $7.473\cdot10^{-12}$& $2.996\cdot10^{-5}$& $0$& $0$\\ 
\bottomrule
\end{tabular}\\
%\textcolor{red}{It is surprising that GLM should be the best model for this data set! Wasn't BRT better in the analyses we did? Can you please check the cross-validation log-likelihoods? For the simulations it makes sense that GLM is excellent, because the true model has the GLM-structure; but not so for the real data sets (see red fox!).} 
%\textcolor{green}{The GLM, improved by stepBIC, is competing here against a rather rudimentary BRT. I have checked all models for mistakes but did not find any.} 
\end{table}

\section{Discussion}

So far, Ye's (\citeyear{Ye1998}) Generalised Degrees of Freedom-concept did not attract much attention in the statistical literature, even though it builds on established principles and applies to machine learning, where model complexity is unknown. \citet{Shen2006} have explored the perturbation approach to GDFs in the context of adaptive model selection for linear models and later extended it to linear mixed effect models \citep{Zhang2012}. The also extended it to the exponential family (including the Bernoulli distribution) and even to classification and regression trees \citep{Shen2004}. Our study differs in that the models considered are more diverse and internally include weighted averaging, which clearly poses a challenge to the GDF-algorithm.

\subsection{GDF for normally distributed data}

For normally distributed data our explorations demonstrate a low sensitivity to the intensity of perturbation used to compute GDFs. Furthermore, across the five modelling approaches employed here, GDF estimates are stable and constant for different numbers of data points perturbed simultaneously.

GDF-estimates were consistent with the rank of GLM models and in line with the estimated degrees of freedom reported by the GAM. For neural networks and boosted regression trees, GDF-estimates appear plausible, but cannot be compared with any self-reported values. Compared to the cross-validation method, GDF values are typically, but not consistently, lower by 10-30\% (Table~\ref{mcomplex}).

For randomForest, GDF-estimates were essentially centred on zero. It seems strange to find that an algorithm that uses hundreds of classification and regression trees internally to actually have no (or even negative) degrees of freedom. We expected a low value due to the averaging of submodels \citep[called the `paradox of ensembles' by][]{ElderIV2003}, but not such a complete insensitivity to perturbations. (Within this study, SH and CFD independently reprogrammed our GDF-function to make sure that this was not due to a programming error.) As eqn.~\ref{gdf_vert} shows, the perturbation of individual data point are compared to the change in the model expectation for this data point, and then summed over all data points. To yield a GDF of 0, the change in expectation (numerator) must be much smaller than the perturbation itself (denominator). This is possible when expectations are variable due to the stochastic nature of the algorithm. It seems that randomForest is much more variable than the other stochastic approaches of boosting and neural networks.

\subsection{Bernoulli GDF}
Changing the value of Bernoulli-data from 0 into 1 (or vice versa) is a stronger perturbation than adding a small amount of normal noise to Gaussian data. As our exploration has shown, the GDF for such Bernoulli-data is indeed much less well-behaved than for the normal data. Not only is the estimated GDF dependent on the number of data points perturbed, also is this dependence different for each modelling approach we used. This makes GDF-computation impractical for Bernoulli data. As a consequence, we did not attempt to extend GDFs in this way to other distributions, as in our perception only a general, distribution- and model-independent algorithm is desirable.

\subsection{GDF vs model complexity from cross-validation}
Cross-validation is typically used to get a non-optimistic assessment of model fit \citep[e.g.][]{Hawkins2004}. As we have shown, it can also be used to compute a measure of model complexity similar (in principle) to GDF (eqn.~\ref{eqn:mcCV} and Table~\ref{mcomplex}). Both express model complexity as the effective number of parameters fitted in the model. GDF and cross-validation-based model complexity estimator $\Delta \ell^\text{CV}_m$ are largely similar, but may also differ substantially (Table~\ref{fig:aic_cv}, red fox case study). Since the `correct' value for this estimator is unknown, we cannot tell which approach actually works better. Given our inability to choose the optimal number of data points to perturb (except for GLM), we prefer $\Delta \ell^\text{CV}_m$, which does not make any such assumption.

\subsection{Remaining problems}

To make the GDF approach more generally applicable, a new approach has to be found. The original idea of \citet{Ye1998} is appealing, but not readily transferable in the way we had hoped.

Another problem, even for Gaussian data where this approach seems to be performing fine, is the high computational burden. GDF-estimation requires tens of thousands of model evaluations, giving it very limited appeal, except for small data sets and fast modelling approaches. Cross-validation, as alternative, is at least an order of magnitude faster, but still requires around 1000 evaluations. If the aim is to compute model weights for model averaging, no \emph{precise} estimation of model complexity is needed and even the results of a single 10-fold cross-validation based on eqn.~\ref{eqn:mcCV} can be used. It was beyond the scope of this study to develop an efficient cross-validation-based approach to compute degrees of freedom, but we clearly see this as a more promising way forward than GDF.

\subsection{Alternatives to AIC}
The selection of the most appropriate statistical model is most commonly based on Kullback-Leibler (KL) discrepancy \citep{Kullback1951}: a measure representing the distance between the true and an approximating model. Thus, we assume that a model $m$, for which the distance to the true model is minimal, is the KL-best model. Yet, since KL-discrepancy is not observable, even if a true model existed, many statisticians have attempted to find a metric approximation \citep[e.g.][]{Burnham2002,Burnham1994}. \citet{Akaike1973}, who proposed this measure as the basis for model selection in the first place, developed the AIC to get around the discussed problem.  

The point of the cross-validated log-likelihood is that we do away with the approximation that yields the degrees of freedom term in the AIC, instead estimating the model-dependent part of the KL divergence directly.
This approach is disadvantageous if AIC can be computed from a single model fit. But if the EDF terms for the AIC would require repeated model fits then there is no reason to use the AIC-approximation to the KL-divergence, rather than a more direct estimator. If leave-one-out cross-validation is too expensive, then we can leave out several, at the cost of some Monte-Carlo variability (resulting from the fact that averaging over all possible left out sets is generally impossible).    

%According to \citet{Burnham2002}, the Akaike weight $w_{\text{AIC}}$ is the actual probability of the respective model $m$ to be the expected KL best model. They refer to Akaike (\citeyear{Akaike1981,Akaike1994}) and \citet{Bozdogan1987}, neither of whom can prove this statement.
       
\section{Conclusion}
%This paper deals with GDF and their adaptation to blackbox algorithms and binary data. So far, Ye's (\citeyear{Ye1998}) concept did not attract much attention in scientific research, even though there is no alternative way to account for component complexity in machine learning models. There might be ways to improve this approach, eventually leading to  accurate results. However, with an already given alternative much more readily used, the benefit of GDF is doubtful. The enormous computing time in combination with the fact that computation is generally a rather difficult undertaking, do not make GDF particularly worthwhile.\\
%As an alternative to GDF-based AIC, cross-validation offers a conveniently computable measurement.          
%Given this situation, one presumably tends to question the AIC framework as the most attractive assessment tool in blackbox model selection or averaging.\\

We have shown that the idea of using GDFs to extend information-theoretical measures of model fit (such as AIC) to non-likelihood models is burdened with large computational costs and yields variable results for different modelling approaches. Cross-validation is more variable than GDFs, but a more direct way to compute measures of model complexity, fit and weights (in a model averaging context). As cross-validation may, but need not, employ the likelihood fit to the hold-out, it appears more plausible for models that do not make likelihood assumptions. Thus, we recommend repeated ($> 100$ times) 10-fold cross-validation to estimate any of the statistics under consideration.

\section*{Acknowledgements}
This work was partly performed on the computational resource bwUniCluster funded by the Ministry of Science, Research and Arts and the Universities of the State of Baden-Württemberg, Germany, within the framework program bwHPC.

%\textcolor{red}{Severin: Ich habe bis Kullback alle Referenzen überarbeiten müssen (vor allem fehlenden Angaben, aber auch großKleinschreibung). Kannst Du bitte die anderen noch übernehmen?} % All done

% ------------------------------------------------------------ %
% Figures
% ------------------------------------------------------------ %
% References
\bibliographystyle{apalike2} 
\bibliography{BSc_lib}

\begin{thebibliography}{}

\bibitem[Akaike, 1973]{Akaike1973}
Akaike, H. (1973).
\newblock Information theory and an extension of the maximum likelihood
  principle.
\newblock In {\em Second International Symposium on Information Theory}  (pp.\
  267--281).: Akademinai Kiado.

\bibitem[Burnham \& Anderson, 1998]{Burnham1998}
Burnham, K.~P. \& Anderson, D.~R. (1998).
\newblock {\em Model Selection and Multimodel Inference: A Practical
  Information-Theoretic Approach}.
\newblock Berlin: Springer.

\bibitem[Burnham \& Anderson, 2002]{Burnham2002}
Burnham, K.~P. \& Anderson, D.~R. (2002).
\newblock {\em Model Selection and Multimodel Inference: A Practical
  Information-Theoretic Approach}.
\newblock Berlin: Springer.
\newblock 2nd edition.

\bibitem[Burnham et~al., 1994]{Burnham1994}
Burnham, K.~P., Anderson, D.~R., \& White, G.~C. (1994).
\newblock Evaluation of the {K}ullback-{L}eibler discrepancy for model
  selection in open population capture-recapture models.
\newblock {\em Biometrical Journal}, 36(3), 299--315.

\bibitem[Diniz-Filho et~al., 2008]{Diniz-Filho2008}
Diniz-Filho, J. A.~F., Rangel, T. F. L. V.~B., \& Bini, L.~M. (2008).
\newblock Model selection and information theory in geographical ecology.
\newblock {\em Global Ecology and Biogeography}, 17, 479--488.

\bibitem[Dormann \& Kaschner, 2010]{Dormann2010}
Dormann, C. \& Kaschner, K. (2010).
\newblock Where's the sperm whale? {A} species distribution example analysis.
\newblock \url{http://www.mced-ecology.org/?page_id=355}.

\bibitem[Dormann et~al., 2010]{Dormann2010a}
Dormann, C.~F., Gruber, B., Winter, M., \& Herrmann, D. (2010).
\newblock Evolution of climate niches in european mammals?
\newblock {\em Biology Letters}, 6, 229--232.

\bibitem[Draper, 1995]{Draper1995}
Draper, D. (1995).
\newblock Assessment and propagation of model uncertainty.
\newblock {\em Journal of the Royal Statistical Society Series B}, 57, 45--97.

\bibitem[Elder, 2003]{ElderIV2003}
Elder, J.~F. (2003).
\newblock The generalization paradox of ensembles.
\newblock {\em Journal of Computational and Graphical Statistics}, 12(4),
  853--864.

\bibitem[Elith et~al., 2006]{Elith2006}
Elith, J., Graham, C.~H., Anderson, R.~P., Dudík, M., Ferrier, S., Guisan, A.,
  Hijmans, R.~J., Huettmann, F., Leathwick, J.~R., Lehmann, A., Li, J.,
  Lohmann, L.~G., Loiselle, B.~A., Manion, G., Moritz, C., Nakamura, M.,
  Nakazawa, Y., Overton, J.~M., Peterson, A.~T., Phillips, S.~J., Richardson,
  K., Scachetti-Pereira, R., Schapire, R.~E., Soberón, J., Williams, S., Wisz,
  M.~S., Zimmermann, N.~E., \& Araujo, M. (2006).
\newblock Novel methods improve prediction of species' distributions from
  occurrence data.
\newblock {\em Ecography}, 29(2), 129--151.

\bibitem[Elith \& Leathwick, 2009]{Elith2009}
Elith, J. \& Leathwick, J.~R. (2009).
\newblock Species distribution models: ecological explanation and prediction
  across space and time.
\newblock {\em Annual Review of Ecology, Evolution, and Systematics}, 40(1),
  677--697.

\bibitem[Hastie et~al., 2009]{Hastie2009}
Hastie, T., Tibshirani, R., \& Friedman, J. (2009).
\newblock {\em The Elements of Statistical Learning: Data Mining, Inference,
  and Prediction}, volume~2.
\newblock Berlin: Springer.

\bibitem[Hawkins, 2004]{Hawkins2004}
Hawkins, D.~M. (2004).
\newblock The problem of overfitting.
\newblock {\em Journal of Chemical Information and Computer Sciences}, 44(1),
  1--12.

\bibitem[Hegyi \& Garamszegi, 2011]{Hegyi2011}
Hegyi, G. \& Garamszegi, L.~Z. (2011).
\newblock Using information theory as a substitute for stepwise regression in
  ecology and behavior.
\newblock {\em Behavioral Ecology and Sociobiology}, 65(1), 69--76.

\bibitem[Horne \& Garton, 2006]{Horne2006}
Horne, J.~S. \& Garton, E.~O. (2006).
\newblock Likelihood cross-validation versus least squares cross-validation for
  choosing the smoothing parameter in kernel home-range analysis.
\newblock {\em Journal of Wildlife Management}, 70, 641--648.

\bibitem[Hurvich \& Tsai, 1989]{Hurvich1989}
Hurvich, C.~M. \& Tsai, C.-L. (1989).
\newblock Regression and time series model selection in small samples.
\newblock {\em Biometrika}, 76(2), 297--307.

\bibitem[Kullback \& Leibler, 1951]{Kullback1951}
Kullback, S. \& Leibler, R.~A. (1951).
\newblock On information and sufficiency.
\newblock {\em The Annals of Mathematical Statistics}, 1, 79--86.

\bibitem[Liaw \& Wiener, 2002]{Liaw2002}
Liaw, A. \& Wiener, M. (2002).
\newblock Classification and regression by random{F}orest.
\newblock {\em R News}, 2(3), 18--22.

\bibitem[Madigan et~al., 1995]{Madigan1995}
Madigan, D., York, J., \& Allard, D. (1995).
\newblock Bayesian graphical models for discrete data.
\newblock {\em International Statistical Review / Revue Internationale de
  Statistique}, 63(2), 215--232.

\bibitem[Mundry, 2011]{Mundry2011}
Mundry, R. (2011).
\newblock Issues in information theory-based statistical inference --
  commentary from a frequentist's perspective.
\newblock {\em Behavioral Ecology and Sociobiology}, 65(1), 57--68.

\bibitem[Olden et~al., 2008]{Olden2008}
Olden, J.~D., Lawler, J.~J., \& Poff, N.~L. (2008).
\newblock Machine learning methods without tears: A primer for ecologists.
\newblock {\em The Quarterly Review of Biology}, 83(2), 171--193.

\bibitem[Raftery, 1996]{Raftery1996}
Raftery, A.~E. (1996).
\newblock Approximate bayes factors and accounting for model uncertainty in
  generalised linear models.
\newblock {\em Biometrika}, 83(2), 251--266.

\bibitem[Recknagel, 2001]{Recknagel2001}
Recknagel, F. (2001).
\newblock Applications of machine learning to ecological modelling.
\newblock {\em Ecological Modelling}, 146(1--3), 303--310.

\bibitem[Ridgeway et~al., 2013]{Ridgeway2013}
Ridgeway, G. et~al. (2013).
\newblock {\em gbm: Generalized Boosted Regression Models}.
\newblock R package version 2.1.

\bibitem[Shen \& Huang, 2006]{Shen2006}
Shen, X. \& Huang, H.-C. (2006).
\newblock Optimal model assessment, selection, and combination.
\newblock {\em Journal of the American Statistical Association}, 101(474),
  554--568.

\bibitem[Shen et~al., 2004]{Shen2004}
Shen, X., Huang, H.-C., \& Ye, J. (2004).
\newblock Adaptive model selection and assessment for exponential family
  distributions.
\newblock {\em Technometrics}, 46(3), 306--317.

\bibitem[Spiegelhalter et~al., 2002]{Spiegelhalter2002}
Spiegelhalter, D.~J., Best, N.~G., Carlin, B.~P., \& van~der Linde, A. (2002).
\newblock Bayesian measures of model complexity and fit.
\newblock {\em Journal of the Royal Statistical Society B}, 64, 583–639.

\bibitem[Stone, 1977]{Stone1977}
Stone, M. (1977).
\newblock An asymptotic equivalence of choice of model by cross-validation and
  akaike's criterion.
\newblock {\em Journal of the Royal Statistical Society. Series B
  (Methodological)}, 39(1), 44--47.

\bibitem[Sugiura, 1978]{Sugiur1978}
Sugiura, N. (1978).
\newblock Further analysts of the data by akaike' s information criterion and
  the finite corrections.
\newblock {\em Communications in Statistics - Theory and Methods}, 7(1),
  13--26.

\bibitem[Team, 2014]{RCT2014}
Team, R.~C. (2014).
\newblock {\em R: A Language and Environment for Statistical Computing}.
\newblock R Foundation for Statistical Computing, Vienna, Austria.

\bibitem[Turkheimer et~al., 2003]{Turkheimer2003}
Turkheimer, F.~E., Hinz, R., \& Cunningham, V.~J. (2003).
\newblock On the undecidability among kinetic models: From model selection to
  model averaging.
\newblock {\em Journal of Cerebral Blood Flow \& Metabolism}, 23(4), 490--498.

\bibitem[Venables \& Ripley, 2002]{Venables2002}
Venables, W.~N. \& Ripley, B.~D. (2002).
\newblock {\em Modern Applied Statistics with S}.
\newblock New York: Springer, 4th edition.

\bibitem[Wood, 2006]{Wood2006}
Wood, S. (2006).
\newblock {\em Generalized Additive Models: An Introduction with R}.
\newblock New York: Chapman and Hall/CRC.

\bibitem[Wood, 2015]{Wood2015}
Wood, S.~N. (2015).
\newblock {\em Core Statistics}.
\newblock Cambridge: Cambridge University Press.

\bibitem[Ye, 1998]{Ye1998}
Ye, J. (1998).
\newblock On measuring and correcting the effects of data mining and model
  selection.
\newblock {\em Journal of the American Statistical Association}, 93(441),
  120--131.

\bibitem[Zhang et~al., 2012]{Zhang2012}
Zhang, B., Shen, X., \& Mumford, S.~L. (2012).
\newblock Generalized degrees of freedom and adaptive model selection in linear
  mixed-effects models.
\newblock {\em Computational Statistics and Data Analysis}, 56(3), 574–586.

\end{thebibliography}
\end{document}